\documentclass[10pt,journal]{IEEEtran}

\ifCLASSOPTIONcompsoc
  \usepackage[nocompress]{cite}
\else
  \usepackage{cite}
\fi

\ifCLASSINFOpdf
   \usepackage[pdftex]{graphicx}
\else
\fi
\usepackage{color,balance}
\usepackage{caption}
\usepackage{subcaption}
\newtheorem{theorem}{Theorem}[section]
\newtheorem{lemma}[theorem]{Lemma}

\usepackage{amsmath}
\usepackage{amssymb}
\usepackage{mathtools}
\newcommand\myeq{\stackrel{\mathclap{\normalfont\mbox{{\tiny def}}}}{=}}

\begin{document}
\title{Discriminative Bayesian Dictionary Learning \\for  Classification}

\author{Naveed~Akhtar,
        Faisal~Shafait,
        and~Ajmal~Mian
\IEEEcompsocitemizethanks{\IEEEcompsocthanksitem N. Akhtar, F. Shafait and A. Mian are with the School 
of Computer Science and Software Engineering, The University of Western Australia, 35 Stirling Highway Crawley, 6009. WA. naveed.akhtar@research.uwa.edu.au, \{faisal.shafait, ajmal.mian\}@uwa.edu.au.}\\ 


}

\maketitle

\IEEEdisplaynontitleabstractindextext
\IEEEpeerreviewmaketitle

\ifCLASSOPTIONcompsoc
\IEEEraisesectionheading{\section{Introduction}\label{sec:introduction}}
\else
\section*{Abstract}

 We propose a Bayesian approach to learn discriminative dictionaries for sparse representation of data.
The proposed approach infers probability distributions over the atoms of a discriminative dictionary using a Beta Process.
It also computes sets of Bernoulli distributions that associate class labels to the learned dictionary atoms. 
This association signifies the selection probabilities of the dictionary atoms in the expansion of class-specific data. 
Furthermore, the non-parametric character of the proposed approach allows it to infer the correct size of the dictionary. 
We exploit the aforementioned Bernoulli distributions in separately learning a linear classifier.  
The classifier uses the same hierarchical Bayesian model as the dictionary, which we present along the analytical inference solution for Gibbs sampling.
For classification, a test instance is first sparsely encoded over the learned dictionary and the codes are fed to the classifier.  
We performed experiments for face and action recognition; and object and scene-category classification using five public datasets and compared the results with state-of-the-art discriminative sparse representation approaches.
Experiments show that the proposed Bayesian approach consistently outperforms the existing approaches.  
\fi

\begin{IEEEkeywords}
Bayesian sparse representation, Discriminative dictionary learning, Supervised learning, Classification.
\end{IEEEkeywords}
\section{Introduction}
\label{sec:introduction}
Sparse representation encodes a signal as a sparse linear combination of redundant basis vectors. 
With its inspirational roots in human vision system \cite{nature}, \cite{vision}, this technique has been successfully employed in image restoration \cite{denoise}, \cite{denoise1}, \cite{denoise2}, compressive sensing \cite{CS}, \cite{CS1} and morphological component analysis~\cite{MCA}.
More recently, sparse representation based approaches have also shown promising results in face recognition and gender classification \cite{LCKSVD}, \cite{SRC}, \cite{FDDL}, \cite{LDL}, \cite{FR1}, \cite{FR2}, \cite{FR3}, texture and handwritten digit classification 
\cite{DGSDL}, \cite{D3}, \cite{D4}, \cite{D5},  natural image and object classification \cite{LCKSVD}, \cite{DLCOPAR}, \cite{SPM}  and human action recognition \cite{A1}, \cite{A2}, \cite{A3}, \cite{A4}.
The success of these approaches comes from the fact that a  sample from a class can generally be well represented as a sparse linear combination of the other samples from the same class, in a lower dimensional manifold~\cite{SRC}. 


For classification, 
a discriminative sparse representation approach first encodes the test instance over a dictionary, i.e. a redundant set of basis vectors, known as atoms.
Therefore, an effective dictionary is critical for the performance of such approaches. 
It is possible to use an off-the-shelf basis (e.g. fast Fourier transform~\cite{FFT} or wavelets~\cite{WLets}) as a generic dictionary to represent data from different domains/classes. 
However, research in the last decade (~\cite{KSVD},  \cite{LCKSVD}, \cite{FDDL},\cite{DLCOPAR}, \cite{denoise}, \cite{BandE},  \cite{BDL}, \cite{JLD}) has provided strong evidence in favor of learning dictionaries using the domain/class-specific training data, especially for   classification and recognition tasks~\cite{FDDL} where class label information of the training data can be exploited in the supervised learning of a dictionary.

Whereas unsupervised dictionary learning approaches (e.g. K-SVD~\cite{KSVD}, Method of Optimal Directions \cite{MOD}) aim at learning faithful signal representations, supervised sparse representation additionally strives for making the dictionaries discriminative. 
For instance, in Sparse Representation based Classification (SRC) scheme, Wright et al.~\cite{SRC} constructed a discriminative dictionary by directly using the training data as the dictionary atoms.
With each atom  associated to a particular class, the query is assigned the label of the class whose associated atoms maximally contribute to the sparse representation of the query.
Impressive results have been achieved for recognition and classification using SRC, however, the computational complexity of this technique becomes prohibitive for large training data.
This has motivated considerable research on learning discriminative dictionaries that would allow sparse representation based classification with much lower computational cost. 

In order to learn a discriminative dictionary, existing  approaches either force subsets of the dictionary atoms to represent data from only specific classes \cite{DLSI}, \cite{FR3}, \cite{Mairal2008} or they associate the complete dictionary to all the classes and constrain their sparse coefficient to be discriminative \cite{DKSVD}, \cite{LCKSVD}, \cite{D2}.
A third category of techniques learns exclusive sets of class specific and common dictionary atoms to separate the common and particular features of the data from different classes  \cite{DLCOPAR}, \cite{IntRelD}.  
Establishing association between the dictionary atoms and the corresponding class labels is a key step of existing methods. However, adaptively building this association is still an open research problem~\cite{LDL}.
Moreover, the strategy of assigning different number of dictionary atoms to different classes and adjusting the overall size of the dictionary become critical for the classification accuracy of the existing approaches, as no principled approach is generally provided to predetermine these parameters.

\begin{figure*}

        \centering
        \begin{subfigure}[b]{0.99\textwidth}
                \includegraphics[width=\textwidth]{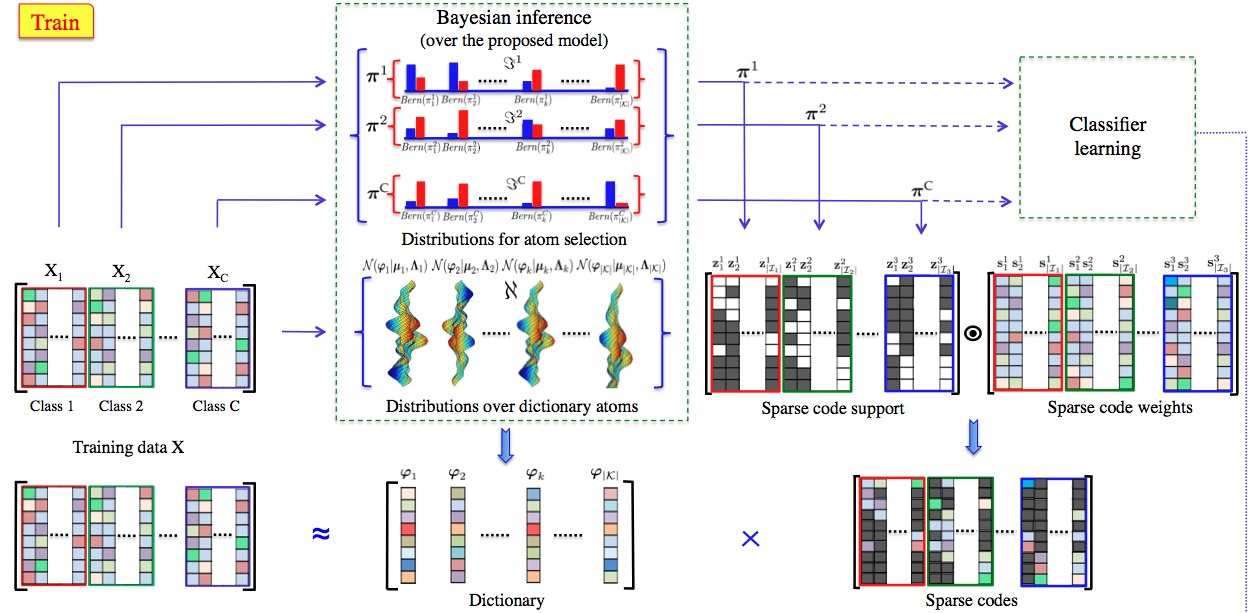}
        \end{subfigure}\\
        \begin{subfigure}[b]{0.99\textwidth}
                \includegraphics[width=\textwidth]{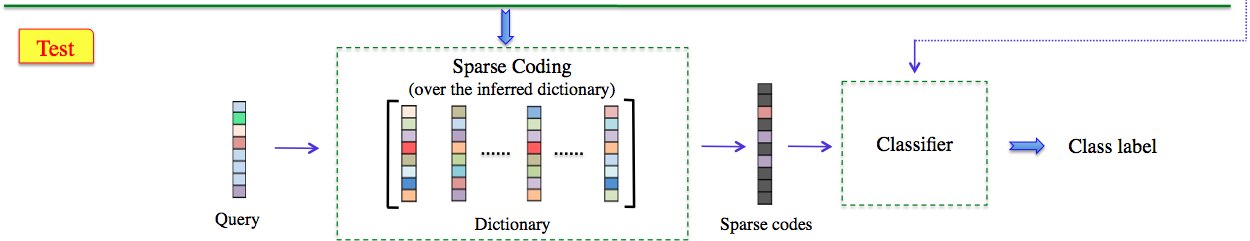}
        \end{subfigure}
        \caption{A schematic diagram of the proposed approach: For training, a set of probability distributions over the dictionary atoms, i.e.~$\mathcal \aleph$,  is learned. We also infer sets of Bernoulli distributions indicating the probabilities of selection of the dictionary atoms in the expansion of data from each class. These distributions are used for inferring the support of the sparse codes. The (parameters of) Bernoulli distributions are later used for learning a classifier. The final dictionary is learned by sampling the distributions in $\mathcal \aleph$, whereas the sparse codes are computed as element-wise product of the support and the weights (inferred by the approach) of the codes. Combined, the dictionary and the codes faithfully represent the training data. For testing, sparse codes of the query over the dictionary are computed and fed to the classifier for labeling.}
        \label{fig:schema}
\end{figure*}

In this work, we propose a solution to this problem by approaching  the sparse representation based classification from a non-parametric Bayesian perspective.
We propose a Bayesian sparse representation technique that infers a discriminative  dictionary using a Beta Process~\cite{BetaP}.
Our approach adaptively builds the association between the dictionary atoms and the class labels such that this association  signifies the probability of selection of the dictionary atoms in the expansion of class-specific data. 
Furthermore, the non-parametric character of the approach allows it to automatically infer the correct size of the dictionary.
The scheme employed by our approach is shown in Fig.~\ref{fig:schema}.
We perform Bayesian inference over a model proposed for  discriminative sparse representation of the training data.
The inference process learns distributions over the dictionary atoms and sets of Bernoulli distributions associating the dictionary  atoms to the labels of the data.  
The Bernoulli distributions govern the support of the final sparse codes and are later utilized in learning a multi-class linear classifier. 
The final dictionary is learned by sampling the distributions over the dictionary atoms and the corresponding sparse codes are computed by element-wise product of the support and the inferred weights of the codes. 
The computed dictionary and the sparse codes also represent the training data faithfully.

A query is classified in our approach by first sparsely encoding it over the inferred dictionary and then classifying its sparse code with the learned classifier. 
In this work, we learn the classifier and the dictionary using the same hierarchical Bayesian model. 
This allows us to exploit the aforementioned Bernoulli distributions in the accurate estimate of the classifier.
We present the proposed Bayesian model along its inference equations for Gibbs sampling.
Our approach has been tested on two face-databases~\cite{AR},~\cite{EYaleB}, an object-database~\cite{C101}, an action-database~\cite{UCF} and a scene-database~\cite{Scene15}.
The classification results are compared with the state-of-the-art discriminative sparse representation approaches.
The proposed approach not only outperforms these approaches in terms of accuracy, its computational efficiency for the classification stage is also comparable to the most efficient existing approaches.



 This paper is organized as follows. 
 We review the related work in Section \ref{sec:RW} of the paper.
In Section \ref{sec:PFnB}, we formulate the problem and briefly explain the relevant concepts that clarify the rationale behind our approach.
The proposed approach is presented in Section \ref{sec:PA},  which includes details of the proposed model, the Gibbs sampling process, the classification scheme and the initialization of the proposed approach. 
Experimental results are reported in Section~\ref{sec:E}  and a discussion on the parameter settings is provided in Section~\ref{sec:D}.
We draw conclusions in Section \ref{sec:Conc}. 



\section{Related Work}
\label{sec:RW}

There are three main categories of the approaches that learn discriminative sparse representation.
In the first category, the learned dictionary atoms have direct correspondence to the labels of the classes  \cite{FR3}, \cite{Mairal2008}, \cite{DLSI}, \cite{Spiro2010}, \cite{A3},  \cite{Wu2010},  \cite{A4}. 
Yang et al. \cite{FR3} proposed an SRC like framework for face recognition, where the atoms of the dictionary are learned from the training data instead of directly using the training data as the dictionary.
In order to learn a dictionary that is simultaneously discriminative and reconstructive, Mairal et al.~\cite{Mairal2008} used a discriminative penalty term in the K-SVD  model \cite{KSVD}, achieving state-of-the-art results on texture segmentation. 
Sprechmann and Sapiro~\cite{Spiro2010} also proposed to learn  dictionaries and sparse codes for clustering. 
In \cite{A4}, Castrodad and Sapiro computed class-specific dictionaries for actions.
The dictionary atoms and their sparse coefficients also exploited the non-negativity of the signals in their approach. 
Active basis models are learned from the training images of each class and applied to object detection and recognition in \cite{Wu2010}.
Ramirez et al.~\cite{DLSI} have used an incoherence promoting term for the dictionary atoms in their learning model. 
Encouraging incoherence among the class-specific sub-dictionaries allowed them to represent samples from the same class better than the samples from the other classes.
Wang et al.~\cite{A3} have proposed to learn class-specific dictionaries for modeling individual actions for action recognition. 
Their model incorporated a similarity constrained term and a dictionary incoherence term for classification.
The above mentioned methods mainly associate a dictionary atom directly to a single class.
Therefore, a query is generally assigned the label of the class whose associated atoms result in the minimum representational error for the query.
The classification stages of the approaches under this category often require the computation of representations of the query over many sub-dictionaries.

In the second category of the approaches for discriminative sparse representation, a single dictionary is shared by all the classes, however the representation coefficients are forced to be discriminative (\cite{LCKSVD}, \cite{D2}, \cite{DKSVD}, \cite{D3}, \cite{D4}, \cite{JLD},  \cite{D5}, \cite{MaxM}, \cite{A1}, \cite{SubMod} ).
Jiang et al.\cite{LCKSVD} proposed a dictionary learning model that encourages the sparse representation coefficients of the same class to be similar.
This is done by adding a 'discriminative sparse-code error'  constraint to a unified objective function that already contains reconstruction error and classification error constraints.   
A similar approach is taken by Rodriguez and Sapiro \cite{D4}  where the authors solve for a simultaneous sparse approximation problem \cite{SOMP} while learning the coefficients. 
It is common to learn dictionaries jointly with a classifier.
Pham and Venkatesh \cite{JLD} and Mairal et al. \cite{D2} proposed to train linear classifiers along the joint dictionaries learned for all the classes.
Zhang and Li \cite{DKSVD} enhanced the K-SVD algorithm~\cite{KSVD} to learn a linear classifier along the dictionary.
A task driven dictionary learning framework has also been proposed \cite{D5}.
Under this framework, different risk functions of the representation coefficients are minimized for different tasks. 
Broadly speaking, the above mentioned approaches aim at learning a single dictionary together with a classifier. 
The query is classified by directly feeding its sparse codes over the learned single dictionary to the classifier.  
Thus, in comparison to the approaches in the first category, the classification stage of these approaches is computationally more efficient.
In terms of learning a single dictionary for the complete training data and the classification stage, the proposed approach also falls under this category of discriminative sparse representation techniques. 

The third category takes a hybrid approach for learning the discriminative sparse representation.
In these approaches, the dictionaries are designed to have a set of shared atoms in addition to class-specific atoms. 
Deng et al.~\cite{ESRC} extended the SRC algorithm by appending an intra-class face variation dictionary to the training data.
This extension achieves promising results in face recognition with a single training sample per class.
Zhou and Fan \cite{IntRelD} employ a Fisher-like regularizer on the representation coefficients while learning a hybrid dictionary. 
Wang and Kong \cite{DLCOPAR} learned a hybrid dictionary to separate the common and particular features of the data.
Their approach additionally encouraged the class-specific dictionaries to be incoherent during the optimization process. 
Shen et al. \cite{MulLevel} proposed to learn a multi-level dictionary for hierarchical visual categorization.
To some extent, it is possible to reduce the size of the dictionary using the hybrid approach, which also results in reducing the classification time in comparison to the approaches that fall under the first category.
However, it is often non-trivial to decide on how to balance between the shared and the class-specific parts of the hybrid dictionary \cite{FDDL}, \cite{LDL}.


\section{Problem Formulation and Background}
\label{sec:PFnB}
Let ${\bf X} = [{\bf X}^1,..., {\bf X}^c,...,{\bf X}^C] \in \mathbb R^{m\times N}$ be the training data comprising $N$ instances from $C$ classes, wherein ${\bf X}^c \in \mathbb R^{m \times N_c}$ represents the data from the $c^{\text {th}}$ class and $\sum_{c = 1}^{C} N_c = N$.
The columns of ${\bf X}^c$ are indexed in $\mathcal I_c$.
We denote a dictionary by $\boldsymbol\Phi \in \mathbb R^{m \times |\mathcal K|}$ with atoms $\boldsymbol\varphi_k$, where $k \in \mathcal K = \{1,...,K\}$ and $|.|$ represents the cardinality of the set. 
Let ${\bf A} \in \mathbb R^{|\mathcal K| \times N}$ be the sparse code matrix of the data, such that ${\bf X} \approx \boldsymbol\Phi {\bf A}$.
We can write ${\bf A} = [{\bf A}^1,...,{\bf A}^c,...,{\bf A}^C]$, where ${\bf A}^c \in \mathbb R^{|\mathcal K| \times |\mathcal I_c|}$ is the sub-matrix related to the $c^{\text{th}}$ class. 
The $i^{\text{th}}$ column of ${\bf A}$ is denoted as $\boldsymbol \alpha_i \in \mathbb R^{|\mathcal K|}$. 
To learn a sparse representation of the data, 
 we can solve the following optimization problem: 
\begin{align}
< \boldsymbol\Phi, {\bf A} > = \min_{\boldsymbol\Phi , {\bf A}} ||{\bf X} - \boldsymbol\Phi {\bf A} ||_F^2 ~~s.t. ~~\forall i, ||\boldsymbol \alpha_i ||_p \leq t, 
\label{eq:gen}
\end{align}
where $t$ is a predefined constant, $||.||_F$ computes the Frobenius norm and $||.||_p$ denotes the $\ell_p$-norm of a vector. 
Generally, $p$ is chosen to be $0$ or $1$ for sparsity~\cite{Book}.
The non-convex optimization problem of Eq.~($\ref{eq:gen}$) can be   iteratively solved by fixing one parameter and solving a convex optimization problem for the other parameter in each iteration.
The solution to Eq.~(\ref{eq:gen}), factors the training data ${\bf X}$ into two complementary matrices, namely the dictionary and the sparse codes,  
without considering the class label information of the training data.
Nevertheless, we can still exploit this factorization in classification tasks by using the sparse codes of the data as features~\cite{LCKSVD}, for which, a classifier can be obtained as
\begin{align}
{\bf W} = \min_{{\bf W}} \sum\limits_{i =1}^N \mathcal L \{h_i, f(\boldsymbol\alpha_i, {\bf W}) \} + \lambda ||{\bf W} ||_F^2, 
\label{eq:classi}
\end{align}
where ${\bf W} \in \mathbb R^{C \times | \mathcal K|}$ contains the model parameters of the classifier, $\mathcal L$ is the loss function, $h_i$ is the label of the $i^{\text{th}}$ training instance ${\bf x}_i \in \mathbb R^m$ and $\lambda$ is the regularization parameter. 

It is usually suboptimal to perform classification based on sparse codes learned by an unsupervised technique.
Considering this, existing approaches \cite{DKSVD}, \cite{JLD},  \cite{D3}, \cite{D2} proposed to jointly optimize a classifier with the dictionary while learning the sparse representation.
One intended ramification of this approach is that the label information also gets induced into the dictionary. 
This happens when the information is utilized in computing the sparse codes of the data, which in turn, are used for computing the dictionary atoms, while solving Eq.~(\ref{eq:gen}).
This results in improving the discriminative abilities of the learned dictionary.
Jiang et al.~\cite{LCKSVD} built further on this concept and  encouraged explicit correspondence between the dictionary atoms and the class-labels.
More precisely, the following optimization problem is solved by the Label-Consistent K-SVD (LC-KSVD2) algorithm~\cite{LCKSVD}:
\begin{align}
<\boldsymbol\Phi, {\bf W}, {\bf T}, {\bf A}> =& \min_{\boldsymbol\Phi, {\bf W}, {\bf T}, {\bf A}}  \Bigg\|  \!\!
\left( \begin{array}{c}
{\bf X}  \\
\sqrt{\upsilon} {\bf Q}  \\
\sqrt{\kappa} {\bf H} \end{array} 
\right)
\!\!-\!\! 
\left( \begin{array}{c}
{\boldsymbol\Phi}  \\
\sqrt{\upsilon} {\bf T}  \\
\sqrt{\kappa} {\bf W} \end{array} 
\right) \!\!{\bf A} \Bigg \|_F^2 \nonumber\\
&s.t. \hspace{1mm} \forall i \hspace{1mm} ||\boldsymbol\alpha_i||_0 \leq t
\label{eq:LCKSVD}
\end{align}  
where $\upsilon$ and $\kappa$ are the regularization parameters, the binary matrix ${\bf H}\in \mathbb R^{C \times N}$ contains the class label information\footnote{For the $i^{\text{th}}$ training instance, the $i^{\text{th}}$ column of ${\bf H}$ has $1$ appearing  only at the index corresponding to the class label.}, ${\bf T} \in \mathbb R^{|\mathcal K| \times |\mathcal K|}$ is the transformation between the sparse codes and the \emph{discriminative sparse codes} ${\bf Q} \in \mathbb R^{|\mathcal K| \times N}$.
Here, for the $i^{\text{th}}$ training instance, the $i^{\text{th}}$ column 
of the fixed binary matrix ${\bf Q}$ has $1$ appearing at the $k^{\text{th}}$ index only if the $k^{\text{th}}$ dictionary atom has the same class label as the training instance. 
Thus, the discriminative sparse codes form a pre-defined relationship between the dictionary atoms and the class labels.
This brings improvement to the discriminative abilities of the dictionary learned by solving Eq. (\ref{eq:LCKSVD}). 

\begin{figure*}
        \centering    
                
        \begin{subfigure}[b]{0.39\textwidth}
                \includegraphics[width=\textwidth]{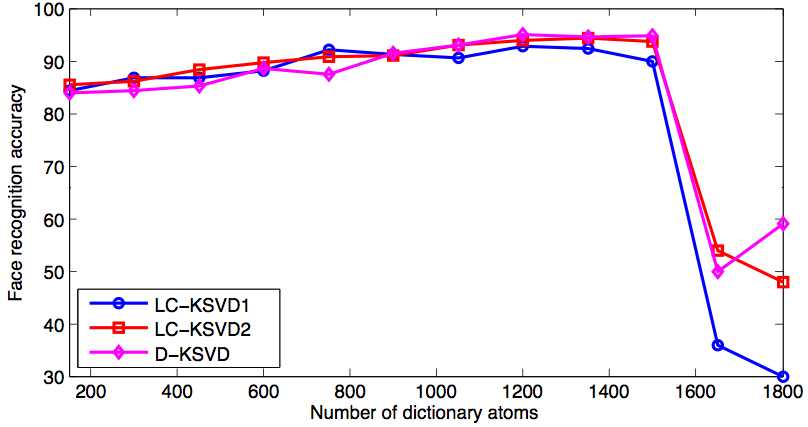}
                \caption{AR database \cite{AR}}
                \label{fig:AR}
        \end{subfigure}
                \hspace{15mm}
        \begin{subfigure}[b]{0.39\textwidth}
                \includegraphics[width=\textwidth]{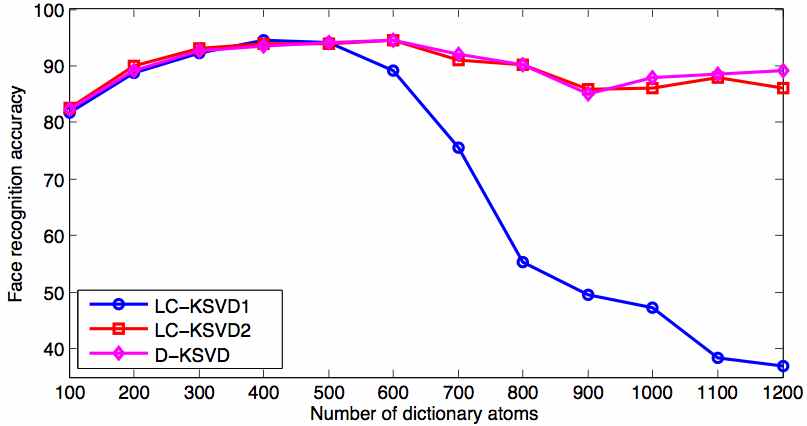}
                \caption{Extended YaleB database~\cite{EYaleB}}
                \label{fir:Yale}
        \end{subfigure}%
        \caption{Examples of how recognition accuracy is affected with varying dictionary size: $\kappa = 0$ for LC-KSVD1 and $\upsilon = 0$ for D-KSVD in Eq. (\ref{eq:LCKSVD}). All other parameters are kept constant at optimal values reported in \cite{LCKSVD}. For the AR database, 2000 training instances are used and testing is performed with 600 instances. For the Extended YaleB, half of the database is used for training and the other half is used for testing. The instances are selected uniformly at random.}
        \label{fig:mok}
\end{figure*}

It is worth noting that in Label-Consistent K-SVD algorithm~\cite{LCKSVD}, the relationship between class-specific subsets of dictionary atoms and class labels is pre-defined.
However, regularization allows flexibility in this association during optimization.
We also note that using $\upsilon = 0$ in Eq. (\ref{eq:LCKSVD}) reduces the optimization problem to the one solved by   Discriminative K-SVD (D-KSVD) algorithm~\cite{DKSVD}.
Successful results are achievable using the above mentioned techniques for recognition and classification.
However, like any discriminative sparse representation approach, these results are obtainable only after careful optimization of the algorithm parameters, including the dictionary size.
In Fig.~\ref{fig:mok}, we illustrate the behavior of recognition accuracy under varying dictionary sizes for \cite{DKSVD} and \cite{LCKSVD} for two face databases. 

Paisley and Carin~\cite{BetaP} developed a \emph{Beta Process} for non-parametric factor analysis, which was later used by Zhou et al. \cite{BDL} 
in successful image restoration and compressive sensing.
Exploiting the non-parametric Bayesian framework, a Beta Process can automatically infer the factor/dictionary size from the training data. 
With the base measure $\hbar_0$ and parameters $a_o > 0$ and $b_o > 0$, a Beta Process is denoted by BP$(a_o, b_o, \hbar_0)$.
A draw from this process, i.e. $\hbar \sim \text{BP} (a_o, b_o, \hbar_0)$,  can be represented as
\begin{align} 
\hbar &= \sum\limits_{k} \pi_k \delta_{\boldsymbol\varphi_k}(\boldsymbol\varphi), \hspace{2.5mm} k \in \mathcal K = \{1,...,K\},\nonumber  \\
\pi_k &\sim \text{Beta}(\pi_k | a_o / K, b_o (K-1)/K ), \nonumber \\
\boldsymbol\varphi_k &\sim \hbar_0,
\label{eq:BP}
\end{align}
with this a valid measure as $K \rightarrow \infty$.
In the above equation, $\delta_{\boldsymbol\varphi_k}(\boldsymbol\varphi)$ is $1$  when $\boldsymbol\varphi = \boldsymbol\varphi_k$ and $0$ otherwise.
Therefore, $\hbar$ can be represented as a set of  $|\mathcal K|$ probabilities, each having an associated vector $\boldsymbol\varphi_k$, drawn $i.i.d.$ from the base measure $\hbar_0$.
Using $\hbar$, we can draw a binary vector ${\bf z}_i \in \{0,1\}^{|\mathcal K|}$, such that the $k^{\text{th}}$ component of this vector is drawn ${ z}_{ik}\sim\text{Bernoulli}(\pi_k)$.
By independently drawing $N$ such vectors, we may construct a matrix ${\bf Z} \in \{0, 1\}^{|\mathcal K| \times N}$, where ${\bf z}_i$ is the $i^{\text{th}}$ column of this matrix.

Using the above mentioned  Beta Process, it is possible to factorize ${\bf X} \in \mathbb R^{m \times N}$ as follows:
\begin{align}
{\bf X} = \boldsymbol\Phi {\bf Z} + {\bf E},
\label{eq:basic}
\end{align} 
where, $\boldsymbol\Phi \in \mathbb R ^{m \times |\mathcal K|}$ has $\boldsymbol\varphi_k$ as its columns and ${\bf E} \in \mathbb R^{m \times N}$ is the error matrix. 
In Eq. (\ref{eq:basic}), the number of non-zero components in a column of ${\bf Z}$ is a random number drawn from Poisson$(a_o/b_o)$ \cite{BetaP}. 
Thus, sparsity can be imposed on the representation with the help of parameters $a_o$ and $b_o$. 
The components of the $k^{\text{th}}$ row of  ${\bf Z}$ 
are independent draws from Bernoulli$(\pi_k)$.
Let $\boldsymbol\pi \in \mathbb R^{|\mathcal K|}$ be a vector with $\pi_{k \in \mathcal K}$, as its $k^{\text{th}}$ element.
This vector governs the probability of selection of the columns of $\boldsymbol \Phi$ in the expansion of the data.
Existence of this physically meaningful latent vector in the Beta Process based matrix factorization plays a central role in the proposed approach for discriminative dictionary learning.



\section{Proposed Approach}
\label{sec:PA}
We propose a Discriminative Bayesian Dictionary Learning  approach for classification.
For the $c^{\text{th}}$ class,
the proposed approach draws $|\mathcal I_c|$ binary vectors ${\bf z}_i^c \in \mathbb R^{|\mathcal K|}$, $\forall i\in \mathcal I_c$ using a Beta Process. 
For each class, the vectors are sampled using separate draws with the same base.
That is, the matrix factorization is governed by a set of $C$ probability vectors $\boldsymbol\pi^{c\in \{1,...,C\}}$, instead of a single vector, however the inferred dictionary is shared by all the classes. 
An element of the aforementioned set, i.e. $\boldsymbol\pi^c \in \mathbb R^{|\mathcal K|}$, controls the probability of selection of the dictionary atoms for a single class data.
This promotes the discriminative abilities of the inferred dictionary.

\subsection{The Model}
\label{sec:M}
Let $\boldsymbol\alpha^c_i \in \mathbb R^{|\mathcal K|}$ denote the sparse code of the $i^{\text{th}}$ training instance of the $c^{\text{th}}$ class, i.e. ${\bf x}_i^c \in \mathbb R^m$, over a dictionary $\boldsymbol\Phi \in \mathbb R^{m \times |\mathcal K|}$.  
Mathematically, ${\bf x}_i^c = \boldsymbol\Phi \boldsymbol\alpha_i^c + \boldsymbol\epsilon_i$, where $\boldsymbol\epsilon_i \in \mathbb R^m$ denotes the modeling error.
We can directly use the Beta Process discussed in Section \ref{sec:PFnB} for computing the desired sparse code and the dictionary.
However, the model employed by the Beta Process is restrictive, as it only allows the code to be binary.
To overcome this restriction, let $\boldsymbol\alpha_i^c = {\bf z}_i^c \odot {\bf s}_i^c$, where $\odot$ denotes the Hadamard/element-wise product, ${\bf z}_i^c \in \mathbb R^{|\mathcal K|}$ is the binary vector and ${\bf s}_i^c \in \mathbb R^{|\mathcal K|}$ is a weight vector.
We place a standard normal prior $\mathcal N( s_{ik}^c | 0, 1/\lambda_{s_o}^{c})$ on the $k^{\text{th}}$ component of the weight vector $s_{ik}^c$, where 
$\lambda_{s_o}^{c}$ denotes the precision of the distribution.
In here, as in the following text, we use the subscript `$o$' to distinguish the parameters of the prior distributions.  
The prior distribution over the $k^{\text{th}}$ component of the binary vector is $\textrm{Bernoulli} ({z}_{ik}^c | \pi_{k_o}^c)$. 
We draw the atoms of the dictionary from a multivariate Gaussian base, i.e. $\boldsymbol\varphi_k  \sim \mathcal N(\boldsymbol\varphi_k | \boldsymbol\mu_{k_o}, \boldsymbol\Lambda^{-1}_{k_o})$, where $\boldsymbol\mu_{k_o} \in \mathbb R^{m}$ is the mean vector and $\boldsymbol\Lambda_{k_o} \in \mathbb R^{m \times m}$ is the precision matrix for the $k^{\text{th}}$ atom of the dictionary. 
We model the error as zero mean Gaussian in $\mathbb R^m$.
Thus, we arrive at the following representation model: 
\begin{align}
{\bf x}_i^c &= \boldsymbol\Phi \boldsymbol\alpha_i^c + \boldsymbol\epsilon_i \hspace{15mm} \forall i \in \mathcal I_c, \forall c \nonumber \\
{\boldsymbol\alpha_i^c} &= {\bf z}_i^c \odot {\bf s}_i^c \nonumber \\
{z}_{ik}^c &\sim \textrm{Bernoulli} ({z}_{ik}^c | \pi_{k_o}^c) \nonumber \\
s_{ik}^c &\sim \mathcal N( s_{ik}^c | 0, 1/\lambda_{s_o}^{c}) \nonumber \\
\pi_{k}^c &\sim \textrm{Beta}(\pi_{k}^c|{a_o}/{K}, {b_o(K-1)}/{K}) \nonumber \\ 
\boldsymbol\varphi_k  &\sim \mathcal N(\boldsymbol\varphi_k | \boldsymbol\mu_{k_o}, \boldsymbol\Lambda^{-1}_{k_o})\hspace{4mm} \forall k \in \mathcal K \nonumber \\
\boldsymbol\epsilon_i &\sim \mathcal N(\boldsymbol\epsilon_i  | {\bf 0}, \boldsymbol\Lambda_{\epsilon_o}^{-1}) \hspace{9mm} \forall i \in \{1,...,N\}
\label{eq:mod}
\end{align}

\begin{figure}[t] 
   \centering
   \includegraphics[width=3in]{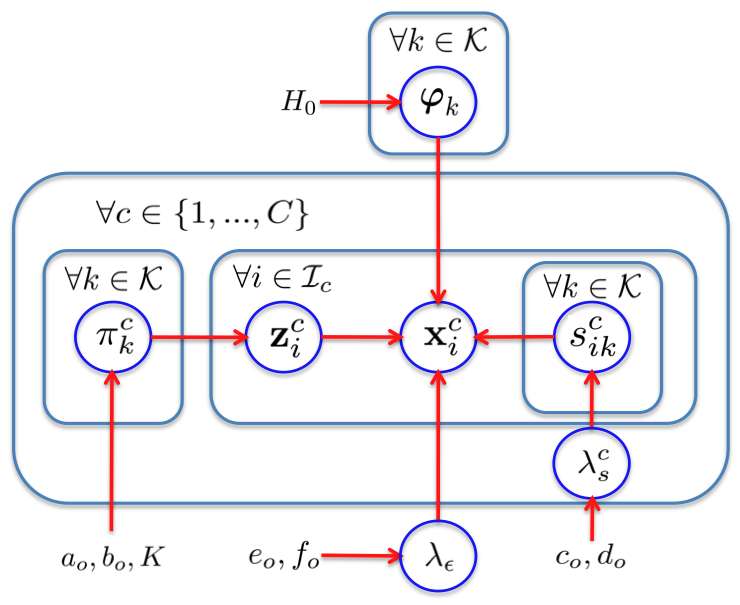} 
   \caption{Graphical representation of the proposed discriminative Bayesian dictionary learning model.} 
   \label{fig:GR}
\end{figure}

Notice, in the above model a conjugate Beta prior is placed over the parameter of the Bernoulli distribution, as mentioned in Section \ref{sec:PFnB}. 
Hence, a latent probability vector $\boldsymbol\pi^c$ (with $\pi_k^c$ as its components) is associated with the dictionary atoms for the representation of the data from the $c^{\text{th}}$ class. 
The common dictionary $\boldsymbol\Phi$ is inferred from $C$ such vectors. 
In the above model, this fact is notationally expressed by showing the dictionary atoms being sampled from a common set of $|\mathcal K|$ distributions, while distinguishing the class-specific variables in the other notations with a superscript `$c$'.   
We assume the same statistics for the modeling error over the complete training data\footnote{It is also possible to use different statistics for different classes, however, in practice the assumption of similar noise statistics works well. We adopt the latter to avoid unnecessary complexity.}.
We further place non-informative Gamma hyper-priors over the precision parameters of the normal distributions. 
That is, $\lambda^c_{s} \sim \Gamma (\lambda^c_{s}|c_o, d_o)$ and  $\lambda_{\epsilon} \sim \Gamma (\lambda_{\epsilon} | e_o, f_o)$, where $c_o, d_o, e_o$ and $f_o$ are the parameters of the respective Gamma distributions.
Here, we allow the error to have an isotropic precision, i.e. $\boldsymbol\Lambda_{\epsilon} = \lambda_{\epsilon} {\bf I}_m$, where ${\bf I}_m$ denotes the identity matrix in $\mathbb R^{m \times m}$. 
The graphical representation of the complete model is shown in Fig.~\ref{fig:GR}. 

\subsection{Inference}
\label{sec:I}
Gibbs sampling is used to perform Bayesian inference over the proposed model\footnote{Paisley and Carin \cite{BetaP} derived variational Bayesian algorithm \cite{VB} for their model. It was shown by Zhou et al. \cite{BDL} that Gibbs sampling is an equally effective strategy in data representation using the same model. Since it is easier to relate the Gibbs sampling process to the learning process of conventional optimization based sparse representation (e.g. K-SVD \cite{KSVD}), we derive expressions for the Gibbs sampler for our approach. Due to the conjugacy of the model, these expressions can be derived analytically.}.
Starting with the dictionary, below we derive analytical expressions for the posterior distributions over the model parameters for the Gibbs sampler.
The inference process performs sampling over these posterior distributions.
The expressions are derived assuming zero mean Gaussian prior over the dictionary atoms, with isotropic precision. 
That is, $\boldsymbol\mu_{k_o} = {\bf 0}$ and $\boldsymbol\Lambda_{k_o} = \lambda_{k_o} {\bf I}_m$. 
This simplification leads to faster sampling, without significantly affecting the accuracy of the approach. 
The sampling process samples the atoms of the dictionary one-by-one from their respective posterior distributions.
This process is analogous to the atom-by-atom dictionary update step of K-SVD~\cite{KSVD}, however the sparse codes remain fixed during our dictionary update.  

{\bf Sampling $\boldsymbol\varphi_k$:}
For our model, we can write the following about the posterior distribution over a dictionary atom:
\begin{align*}
p(\boldsymbol\varphi_k | -)  \propto \prod\limits_{i = 1}^{N}  \mathcal N( {\bf x}_i | \boldsymbol\Phi({\bf z}_i \odot {\bf s}_i), \lambda^{-1}_{\epsilon_o} {\bf I}_m) \mathcal N(\boldsymbol\varphi_k | {\bf 0}, \lambda_{k_o}^{-1} {\bf I}_m).
\end{align*}
Here, we intentionally dropped the superscript `$c$' as the dictionary is updated using the complete training data. 
Let ${\bf x}_{i_{\varphi_k}}\!$ denote the contribution of the dictionary atom $\boldsymbol\varphi_k$ to the $i^{\text{th}}$ training instance ${\bf x}_i$: 
\begin{align}
{\bf x}_{i_{\varphi_k}} = {\bf x}_i - \boldsymbol\Phi({\bf z}_i \odot {\bf s}_i) + \boldsymbol\varphi_k  (z_{ik} \odot  s_{ik}).
\label{eq:atomic}
\end{align}
Using Eq. (\ref{eq:atomic}), we can re-write the aforementioned  proportionality as
\begin{align*}
p(\boldsymbol\varphi_k | -)  \propto  \prod\limits_{i = 1}^{N}  \mathcal N( {\bf x}_{i_{\varphi_k}} | \boldsymbol\varphi_k(z_{ik} s_{ik}), \lambda^{-1}_{\epsilon_o} {\bf I}_m) \mathcal N(\boldsymbol\varphi_k | {\bf 0}, \lambda_{k_o}^{-1} {\bf I}_m).
\end{align*}
Considering the above expression, 
the posterior distribution over a dictionary atom can be written as
\begin{align}
p(\boldsymbol\varphi_k | -)  = \mathcal N (\boldsymbol\varphi_k | \boldsymbol\mu_k, \lambda_k^{-1}{\bf I}_m),
\end{align}
where,
\begin{align*}
\boldsymbol\mu_k &= \frac{\lambda_{\epsilon_o}}{\lambda_k} \sum\limits_{i = 1}^{N}(z_{ik}. s_{ik}) {\bf x}_{i_{\varphi_k}}, \\
 \lambda_k &= \lambda_{k_o} + \lambda_{\epsilon_o} \sum\limits_{i=1}^{N} (z_{ik}. s_{ik})^2. 
\end{align*}

{\bf Sampling $z_{ik}^c$:}
Once the dictionary atoms have been sampled, we sample 
$z_{ik}^c$, $\forall i \in \mathcal {I}_c$, $\forall k \in \mathcal K$.
Using the contribution of the $k^{\text{th}}$ dictionary atom, the posterior probability distribution over $z_{ik}^c$ can be expressed as
\begin{align*}
p(z_{ik}^c| -) \propto \mathcal N( {\bf x}^c_{i_{\varphi_k}} | \boldsymbol\varphi_k(z_{ik}^c.s_{ik}^c), \lambda^{-1}_{\epsilon_o} {\bf I}_m) \text{Bernoulli}(z_{ik}^c| \pi_{k_o}^c).
\end{align*}
Here we are concerned with the $c^{\text{th}}$ class only, therefore ${\bf x}^c_{i_{\varphi_k}}$ is computed with the $c^{\text{th}}$ class data in Eq.~(\ref{eq:atomic}).
With the prior probability of $z_{ik}^c = 1$ given by $\pi_{k_o}^c$, we can write the following about its posterior probability:
\begin{align*}
p(z_{ik}^c = 1| -)\propto \pi_{k_o}^c   \exp{ \left( - \frac{\lambda_{\epsilon_o}}{ 2} ||{\bf x}^c_{i_{\varphi_k}} - \boldsymbol\varphi_k s_{ik}^c||_2^2  \right)}.\
\end{align*}
It can be shown that the right hand side of the above proportionality can be written as:
\begin{align*}
p_1 = \pi_{k_o}^c  \zeta_1 \zeta_2,
\end{align*}
where, $\zeta_1 = \exp\left( - \frac{\lambda_{\epsilon_o} s_{ik}^c }{2}(||\boldsymbol\varphi_k||_2^2  s_{ik}^c  - 2 ({\bf x}_{i_{\varphi_k}}^c)^{\text T}\boldsymbol\varphi_k) \right)$ and $\zeta_2 = \exp\left( - \frac{\lambda_{\epsilon_o}}{2} ||{\bf x}^c_{i_{\varphi_k}}||_2^2\right)$. 
Furthermore, since the prior probability of $z_{ik}^c = 0$ is given by $1-\pi_{k_o}^c$, we can write the following about its posterior probability:
\begin{align*}
p(z_{ik}^c = 0| -) &\propto (1-\pi_{k_o}^c   ) \zeta_2.
\end{align*}
Thus, $z_{ik}^c$ can be sampled from the following normalized Bernoulli distribution: 
\begin{align*}
 \text{Bernoulli}\left( z_{ik}^c \Big | \frac{p_1}{p_1+ (1-\pi_{k_o}^c)\zeta_2}\right).
\end{align*}
By inserting the value of $p_1$ and simplifying, we finally arrive at the following expression for sampling $z_{ik}^c$:
\begin{align}
z_{ik}^c \sim \text{Bernoulli}\left( z_{ik}^c \Big | \frac{\pi_{k_o}^c \zeta_1 }{1+ \pi_{k_o}^c(\zeta_1 - 1)}\right).
\label{eq:berni}
\end{align}

{\bf Sampling $s_{ik}^c$:}
We can write the following about the posterior distribution over $s_{ik}^c$:
\begin{align*}
p(s_{ik}^c | -) \propto  \mathcal N( {\bf x}^c_{i_{\varphi_k}} | \boldsymbol\varphi_k(z_{ik}^c.s_{ik}^c), \lambda^{-1}_{\epsilon_o} {\bf I}_m) \mathcal N(s_{ik}^c | 0, 1/\lambda_{s_o}^c).
\end{align*}
Again, notice that we are concerned with the $c^{\text{th}}$ class data only. 
In light of the above expression, 
$s_{ik}^c$ can be sampled from the following posterior distribution:
\begin{align}
p(s_{ik}^c | -)  = \mathcal N (s_{ik}^c | \mu_s^c, 1/ \lambda_s^c),
\end{align}
where,
\begin{align*}
\mu_s^c 
&= \frac{\lambda_{\epsilon_o}} {\lambda_s^c} z_{ik}^c \boldsymbol\varphi_k^{\text T} {\bf x}^c_{i_{\varphi_k}}, \\
\lambda_s^c 
&= \lambda_{s_o}^c +  \lambda_{\epsilon_o} (z_{ik}^c)^2  ||\boldsymbol\varphi_k||_2^2.
\end{align*}

{\bf Sampling $\pi_k^c$:}
 Based on our model, we can also write the posterior probability distribution over $\pi_k^c$ as
 \begin{align*}
 p(\pi_k^c | -) &\!\propto\!  \prod\limits_{i \in \mathcal I_c}\! \text{Bernoulli}(z_{ik}^c | \pi_{k_o}^c) \text{Beta}\left(\pi_{k_o}^c \Big |\frac{{a}_o}{K}, \frac{b_o(K-1)}{K}\right).
  \end{align*}
Using the conjugacy between the distributions, it can be easily shown that the $k^{\text{th}}$ component of $\boldsymbol\pi^c$ must be drawn from the following posterior distribution during the sampling process:
\begin{align}
 p(\pi_k^c | -) \!=\!  \text{Beta} \! \left(\!\pi_k^c \Big|\frac{a_o}{K} + \!\!\sum\limits_{i \in \mathcal I_c} z_{ik}^c , \frac{b_o (K-1)}{K} + |\mathcal I_c| - \!\!\sum\limits_{i \in \mathcal I_c} z_{ik}^c \!\right)
 \label{eq:beta}
\end{align}
  
 {\bf Sampling $\lambda_s^c$:} 
In our model, the components of the weight vectors are drawn from a standard normal distribution.
For a given weight vector, common priors are assumed over the precision parameters of these distributions.
This allows us to express the likelihood function for $\lambda_s^c$ in terms of standard multivariate Gaussian with isotropic precision. 
Thus, we can write the posterior over $\lambda_s^c$ as the following:
 \begin{align*}
 p(&\lambda_s^c | -) \propto \prod\limits_{i \in \mathcal I_c} \mathcal N \left({\bf s_i^c} \Big| {\bf 0}, \frac{1}{\lambda_{s_o}^c}{\bf I}_{|\mathcal K|}\right) \Gamma(\lambda_{s_o}^c| c_o,d_o).\\
 \end{align*}
 Using the conjugacy between the Gaussian and Gamma distributions, it can be shown that $\lambda_s^c$ must be sampled as follows:
 \begin{align}
 \lambda_s^c \sim \Gamma \left(\lambda_s^c \Big | \frac{|\mathcal K| N_c}{2} + c_o , \frac{1}{2}\sum\limits_{i \in \mathcal I_c} ||{\bf s}_i^c||_2^2+ d_o \right).
 \label{eq:12}
 \end{align}

{\bf Sampling $\lambda_\epsilon$:}
We can write the posterior over $\lambda_\epsilon$ as
\begin{align*}
p(\lambda_\epsilon | -) \propto \prod\limits_{i=1}^{N} \mathcal N ({\bf x}_i | \boldsymbol\Phi({\bf z}_i \odot {\bf s}_i), \lambda_{\epsilon_o}^{-1}{\bf I}_m) \Gamma (\lambda_{\epsilon_o} | e_o, f_o).
\end{align*}
Similar to $\lambda_s^c$, we can arrive at the following for sampling $\lambda_\epsilon$ during the inferencing process:
\begin{align}
\lambda_\epsilon \sim \Gamma(\frac{m N}{2} + e_o , \frac{1}{2}\sum\limits_{i=1}^{N} ||{\bf x}_i - \boldsymbol\Phi({\bf z}_i \odot {\bf s}_i)||_2^2+ f_o).
\label{eq:13}
\end{align}

As a result of Bayesian inference over the model, we obtain sets of  posterior distributions over the model parameters.
We are particularly interested in two of them. Namely, the set of distributions over the dictionary atoms $\aleph \myeq \{ \mathcal N (\boldsymbol\varphi_k | \boldsymbol\mu_k, \boldsymbol\Lambda^{-1}_k) : k \in \mathcal K\} \subset \mathbb R^m$, and the set of probability distributions characterized by the vectors $\boldsymbol\pi^{c\in \{1,...,C\}} \in \mathbb R^{|\mathcal K|}$.  
Momentarily, we defer the discussion on the latter. 
The former is used to compute the desired dictionary $\boldsymbol\Phi$.
This is done by drawing multiple samples from the elements of $\aleph$ and estimating the corresponding dictionary atoms as respective means of the samples.
Indeed, the mean parameters of the elements of $\aleph$ can also be chosen as the desired dictionary atoms. 
However, we prefer the former approach  for robustness.

The proposed model and the sampling process also results in inferring the correct size of the desired dictionary.
We present the following Lemmas in this regard: 
\begin{lemma}
\label{lem:1}
For  $K \rightarrow \infty$, $\mathbb E [\xi] = \frac{a_o}{b_o},\forall c$, where $\xi = \sum\limits_{k = 1}^{K} z_{ik}^c$.  
\end{lemma}
\
\\
\
{\bf Proof:}\footnote{We follow \cite{BetaP} closely in the proof, however, our analysis also takes into account the class labels of the data, whereas no such data discrimination is assumed in \cite{BetaP}.}
According to the proposed model, the covariance of a data vector from the $c^{\text{th}}$ class can be given by:
\begin{align}
\mathbb E[({\bf x}_i^c) ({\bf x}_i^c)^{\text T}] =  \frac{a_o K}{ a_o + b_o(K-1)}  \frac{ \boldsymbol\Lambda_{k_o}^{-1}} {\lambda_{s_o}^c} + \boldsymbol\Lambda_{\epsilon_o}^{-1}
\label{eq:cov}
\end{align}
In Eq.~(\ref{eq:cov}), fraction $\frac{a_o}{a_o + b_o(K-1)}$ appears due to the presence of ${\bf z_i^c}$ in the model and the equation simplifies to $ \mathbb E[({\bf x}_i^c) ({\bf x}_i^c)^{\text T}]  = K \frac{ \boldsymbol\Lambda_{k_o}^{-1}} {\lambda_{s_o}^c} + \boldsymbol\Lambda_{\epsilon_o}^{-1}$ when we neglect ${\bf z}_i^c$.
Here, $K$ signifies the number of dictionary atoms required to  represent the data vector.
Notice in the equation, that as $K \rightarrow \infty$, we observe $\mathbb E[({\bf x}_i^c) ({\bf x}_i^c)^{\text T}] \rightarrow \frac{a_o}{b_o} \frac{\boldsymbol\Lambda_{k_o}^{-1}}{\lambda_{s_o}^c} + \boldsymbol\Lambda^{-1}_{\epsilon_o}$. 
Thus, in the limit $K \rightarrow \infty$, $\frac{a_o}{b_o}$ corresponds to the expected number of non-zero components in ${\bf z}_i^c$, given by $\mathbb E[\xi]$, where $\xi = \sum\limits_{k = 1}^{K} z_{ik}^c$.

\begin{lemma}
\label{lem:2}
Once $\pi_k^c = 0$ in a given iteration of the sampling process, $\mathbb E [\pi_k^c] \approx 0$ for the later iterations. 
\end{lemma}
\
\\
\
{\bf Proof:}
According to Eq.~(\ref{eq:berni}),  $\forall i \in \mathcal I_c$, $z_{ik}^c = 0$ when $\pi_{k_o}^c = 0$. 
Once this happens, the posterior distribution over $\pi_{k}^c$ becomes $\text{Beta} \left(\pi_k^c \Big| \hat a, \hat b \right)$, where $\hat a = \frac{a_o}{K}$ and  $\hat b = \frac{b_o (K-1)}{K} + |\mathcal I_c|$ (see Eq. \ref{eq:beta}).
Thus, the expected value of $\pi_k^c$ for the later iterations can be written as $\mathbb E[\pi_k^c] = \frac{\hat a}{ \hat a + \hat b} = \frac{a_o}{ a_o + b_o(K-1) + K |\mathcal I_c|}$.
With $0 < a_o,b_o < |\mathcal I_c| \ll K$ we can see that $\mathbb E [\pi_k^c] \approx 0$. 
 
In the Gibbs sampling process, we start with $K  \rightarrow \infty$ in our implementation and let $0 < a_o,b_o < |\mathcal I_c|$.
Considering Lemma~\ref{lem:1}, the values of $a_o$ and $b_o$ are set to ensure that the resulting representation is sparse. 
We drop the $k^{\text{th}}$ dictionary atom during the sampling  process if $\pi_{k}^c = 0$, for all the classes simultaneously. 
According to Lemma~\ref{lem:2}, dropping such an atom does not bring significant changes to the final representation.
Thus, by removing the redundant dictionary atoms in each sampling iteration, we finally arrive at the correct size of the desired dictionary, i.e. $|\mathcal K|$.

\begin{figure*}[t] 
   \centering
   \includegraphics[width=\textwidth]{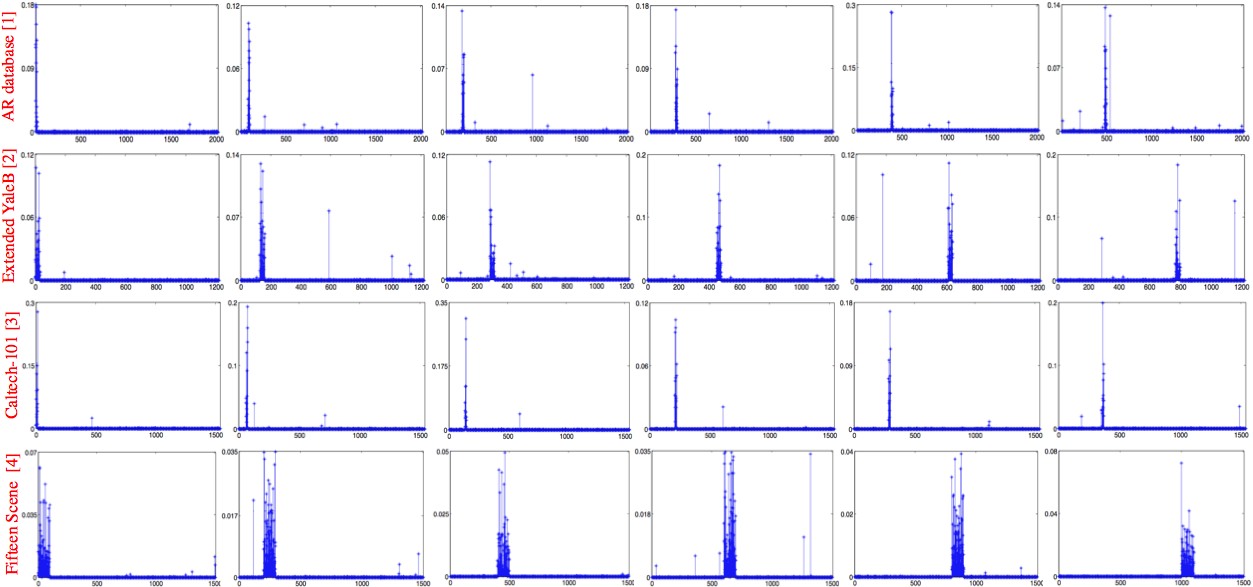} 
   \caption{Illustration of the discriminative character of the inferred  dictionary: From top, the four rows present results on AR database \cite{AR}, Extended YaleB \cite{EYaleB}, Caltech-101\cite{C101} and Fifteen Scene categories \cite{Scene15}, respectively. In each plot, the x-axis represents $k \in \mathcal K$ and the y-axis shows the corresponding probability of selection of the $k^{\text{th}}$ dictionary atom in the expansion of the data. A plot represents a single $\boldsymbol\pi^c$ vector learned as a result of Bayesian inference. For the first three rows, from left to right, the value of c (i.e. class label) is 1, 5, 10, 15, 20 and 25, respectively. For the fourth row the value of c is 1, 3, 5, 7, 9 and 11 for the plots from left to right. Plots clearly show distinct clusters of high probabilities for different classes.} 
   \label{fig:probs}
\end{figure*}

As mentioned above, with Bayesian inference over the proposed model we also infer a set of probability vectors $\boldsymbol\pi^{c\in \{1,...,C\}}$. 
Each element of this set, i.e. $\boldsymbol\pi^c \in \mathbb R^{|\mathcal K|}$, further characterizes a set of probability distributions $\Im^c \myeq \{ \text{Bernoulli}(\pi_k^c) : k \in \mathcal K\} \subset \mathbb R$.
Here, $\text{Bernoulli}(\pi_k^c)$ is jointly followed by all the $k^{\text{th}}$ components of the sparse codes for the $c^{\text{th}}$ class.
If the $k^{\text{th}}$ dictionary atom is commonly used in representing the $c^{\text{th}}$ class training data, we must expect a high value of $\pi_k^c$, and $\pi_k^c \rightarrow 0$ otherwise. 
In other words, for an arranged dictionary, components of $\boldsymbol\pi^c$ having large values should generally cluster well if the learned dictionary is discriminative. 
Furthermore, these clusters must appear at different locations in the inferred vectors for different classes. 
Such clusterings 
would demonstrate the discriminative character of the inferred dictionary. 
Fig. \ref{fig:probs} verifies this character for the dictionaries inferred under the proposed model.
Each row of the figure plots six different probability vectors (i.e. $\boldsymbol\pi^c$) for different training datasets. 
A clear clustering of the high value components of the vectors is visible in each plot. 
Detailed experiments are  presented in Section~\ref{sec:E}. 

\subsection{Classification}
\label{sec:C}


Let ${\bf y} \in \mathbb R^m$ be a query signal. 
We follow the common methodology \cite{LCKSVD}, \cite{DKSVD} for classification that first encodes ${\bf y}$ over the inferred dictionary such that ${\bf y} = \boldsymbol\Phi \widehat{\boldsymbol\alpha} + \boldsymbol\epsilon$, and then computes ${\boldsymbol\ell} = {\bf W} \widehat{\boldsymbol\alpha}$, where ${\bf W} \in \mathbb R^{C \times |\mathcal K|}$ contains model parameters of a multi-class linear classifier. 
The query is assigned the class label corresponding to the largest component of $\boldsymbol\ell \in \mathbb R^C$. 
The main difference between the classification approach of this work and that of the existing techniques is in the learning process of ${\bf W}$.  
Whereas discrimination is induced in $\boldsymbol\Phi$ by the joint optimization of ${\bf W}$ and $\boldsymbol\Phi$ in the existing techniques (see Eq. \ref{eq:LCKSVD}), this is already achieved in the inference process of the proposed approach. 
Thus, it is possible to optimize a classifier separately from the dictionary learning process without affecting the discriminative abilities of the learned dictionary.

Let ${\bf h}_i^c \in \mathbb R^C$ be a binary vector with the only $1$ appearing at the $c^{\text{th}}$ index, indicating the class of the  training instance ${\bf x}_i^c$.
Let ${\bf H} \in \mathbb R^{C \times N}$ be the binary index matrix formed by such vectors for the complete training data ${\bf X}$.  
We aim at computing ${\bf W}$ such that ${\bf H} = {\bf W} {\bf B} + {\bf E}$, where ${\bf E} \in \mathbb R^{C \times N}$ denotes the modeling error and ${\bf B} \in \mathbb R^{|\mathcal K| \times N}$ is the coefficient matrix.  
Notice that, we can directly use the model in Eq.~(\ref{eq:mod}) to compute ${\bf W}$.
For that, we can write ${\bf h}_i^c = {\bf W} \boldsymbol\beta_i^c + \boldsymbol\epsilon_i$, where $\boldsymbol\beta_i^c \in \mathbb R^{|\mathcal K|}$ is a column of ${\bf B}$. 
Thus, we infer ${\bf W}$ under the Bayesian framework using the  model proposed in Eq.~(\ref{eq:mod}).
While learning this matrix, we perform Gibbs sampling such that the probability vectors $\boldsymbol\pi^{c \in \{1,...,C\}}$ are kept constant to those finally inferred by the dictionary learning stage.
That is, wherever required, the value of $\pi_k^c$ is directly used from $\boldsymbol\pi^{c \in \{1,...,C\}}$ instead of inferring a new value during the sampling process.

The reason for using the same $\boldsymbol\pi^{c \in \{1,...,C\}}$ vectors for inferring ${\bf W}$ and $\boldsymbol\Phi$ is straightforward. 
Since we first sparse code the query over the learned discriminative dictionary, we expect the underlying support of the learned codes to follow some $\boldsymbol\pi^c$ closely. 
Thus, ${\bf W}$ can be expected to classify the learned codes better if the discriminative information regarding their support is encoded in it.
Notice that, unlike the existing approaches (e.g. \cite{DKSVD}, \cite{LCKSVD}) the coupling between ${\bf W}$ and $\boldsymbol\Phi$ is kept probabilistic in our approach. 
We do not assume that the 'exact values' of the sparse codes of the query would match to those of the training sample (and hence ${\bf W}$ and $\boldsymbol\Phi$ should be trained jointly), rather, our assumption is that samples from the same class are more likely explainable using similar basis. 
Therefore, coupling between ${\bf W}$ and $\boldsymbol\Phi$ is kept in terms of probabilistic selection of their columns. 
Our view point also makes Orthogonal Matching Pursuit (OMP)~\cite{OMP} a natural choice for sparse coding the query over the dictionary.
This greedy pursuit algorithm efficiently searches for the right basis to represent the data.
Therefore, we used OMP in sparse coding the query over the learned dictionary. 
 





\subsection{Initialization}
\label{sec:Init}
For inferring the dictionary, we need to first initialize $\boldsymbol\Phi$, ${\bf z}_i^c$, ${\bf s}_i^c$ and $\pi_k^c$.
We initialize $\boldsymbol\Phi$ by randomly selecting the training instances with replacement. 
We sparsely encode ${\bf x}_i^c$ over the initial dictionary using OMP~\cite{OMP}.
The sparse codes are considered as the initial ${\bf s}_i^c$, whereas their support forms the initial vector ${\bf z}_i^c$. 
Computing the initial ${\bf s}_i^c$ and ${\bf z}_i^c$ with other methods, such as regularized least squares, is equally effective. 
We set $\pi_k^c = 0.5, \forall c, \forall k$ for the initialization. 
Notice, this means that all the dictionary atoms initially have equal chances of getting selected in the expansion of a training instance from any class. 
The values of $\pi_k^c, \forall c, \forall k$ finally inferred by the dictionary learning  process serve as the initial values of these parameters for learning the classifier.
Similarly, the vectors ${\bf z}_i^c$ and ${\bf s}_i^c$ computed by the dictionary learning stage are used for initializing the corresponding vectors for the classifier.
We initialize ${\bf W}$ using the ridge regression model \cite{RidgeReg} with the $\ell_2$-norm regularizer and quadratic loss:
\begin{align}
{\bf W} = \min\limits_{{W}} ||{\bf H} - {\bf W} \boldsymbol\alpha_i ||^2 + \lambda ||{\bf W} ||_2^2, \hspace{1mm} \forall i \in \{1,...,N\},
\label{eq:W}
\end{align}
where $\lambda$ is the regularization constant. The computation is done over the complete training data, therefore the superscript `$c$' is dropped in the above equation.




\section{Experiments}
\label{sec:E}
We have evaluated the proposed approach on two face data sets: the Extended YaleB~\cite{EYaleB} and the AR database~\cite{AR}, a data set for object categories: Caltech-101~\cite{C101}, a data set for scene categorization: Fifteen scene categories~\cite{Scene15}, and an action data set: UCF sports actions~\cite{UCF}.  
These data sets are commonly used in the literature for evaluation of sparse representation based classification techniques.
We compare the performance of the proposed approach with SRC~\cite{SRC}, the two variants of Label-Consistent K-SVD~\cite{LCKSVD} (i.e. LC-KSVD1, LC-KSVD2), the Discriminative K-SVD algorithm (D-KSVD)~\cite{DKSVD}, the Fisher Discrimination Dictionary Learning algorithm (FDDL)~\cite{FDDL} and the Dictionary Learning based on separating the Commonalities and the Particularities of the data (DL-COPAR)~\cite{DLCOPAR}.
In our comparisons, we also include results of unsupervised sparse representation based classification that uses K-SVD~\cite{KSVD} as the dictionary learning technique and separately computes a multi-class linear classifier using Eq.~(\ref{eq:W}).

For all of the above mentioned methods, except SRC and D-KSVD, we acquired the public codes from the original authors.
To implement SRC, we used the LASSO~\cite{LASSO} solver of the SPAMS toolbox \cite{SPAMS}.
For D-KSVD, we used the public code provided by Jiang et al.~\cite{LCKSVD} for LC-KSVD2 algorithm  and solved Eq.~(\ref{eq:LCKSVD}) with $\upsilon = 0$.
The experiments are performed on an Intel Core i7-2600 CPU at 3.4 GHz with 8 GB RAM.  
We performed our own experiments using the above mentioned methods and the proposed approach using the same data. 
The parameters of the existing approaches were carefully optimized following the guidelines of the original works.
We mention the used parameter values and, where it exists, we note the difference between our values and those used in the original works. 
In our experiments, these differences were made to favor the existing approaches.
Results of the approaches other than those mentioned above, are taken directly from the literature, where the same experimental protocol has been followed.

For the proposed approach, the used parameter values were as follows.  
In all experiments, we chose $K = 1.5 N$ for initialization, whereas $c_o, d_o, e_o$ and $f_o$ were all set to $10^{-6}$. 
We selected $a_o = b_o = \frac{\min_c |\mathcal I_c|}{2}$, whereas  $ \lambda_{s_o}$ and $\lambda_{k_o}$ were set to $1$ and $m$, respectively. Furthermore, $\lambda_{\epsilon_o}$ was set to $10^6$ for all the datasets except for Fifteen Scene Categories~\cite{Scene15}, where we used $\lambda_{\epsilon_o} = 10^9$.
In each experiment, the Bayesian inference was performed with  35 Gibbs sampling iterations.
We defer further discussion on the selection of the parameter values to Section~\ref{sec:D}.

\subsection{Extended YaleB}
\label{sec:EYB}
\begin{figure}[t]
        \centering
        \begin{subfigure}[b]{0.3\textwidth}
                \includegraphics[width=\textwidth]{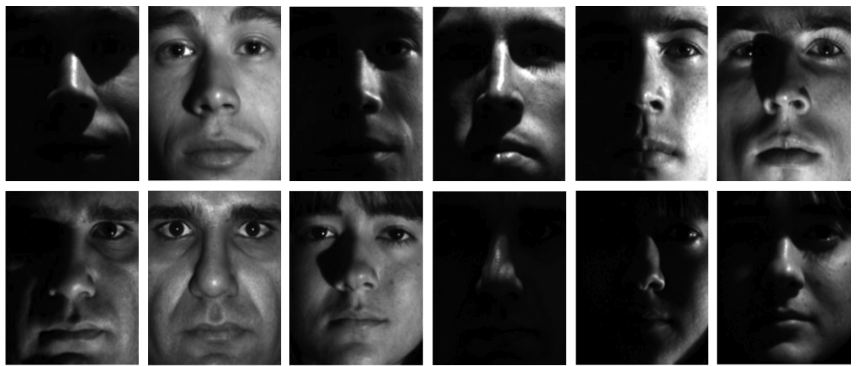}
                \caption{Extended YaleB \cite{EYaleB}}
                \label{fig:Yaleface}
        \end{subfigure}\\%
        \begin{subfigure}[b]{0.3\textwidth}
                \includegraphics[width=\textwidth]{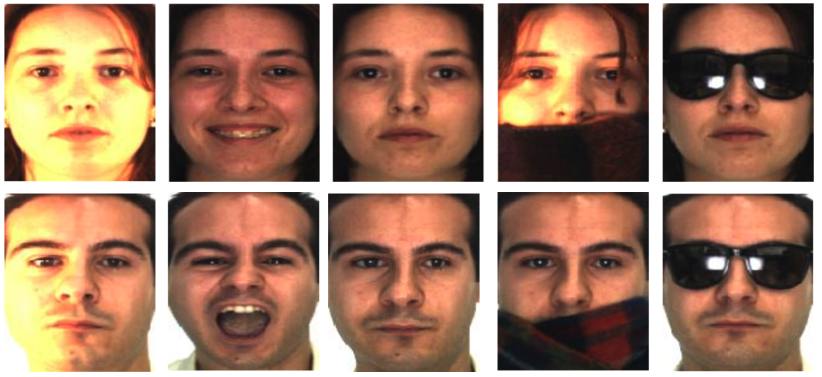}
                \caption{AR database \cite{AR}}
                \label{fig:ARface}
        \end{subfigure}
        \caption{Examples from the face databases.}
        \label{fig:faces}
\end{figure}

Extended YaleB \cite{EYaleB} contains 2,414 frontal face images  of 38 different people, each having about 64 samples. 
The images are acquired under varying illumination conditions and the subjects have different facial expressions. 
This makes the database fairly challenging, see Fig~\ref{fig:Yaleface} for examples.
In our experiments, we used the random face feature descriptor \cite{SRC}, where a cropped $192 \times 168$ pixels image was projected onto a 504-dimensional vector.
For this, the projection matrix was generated from random samples of standard normal distributions. 
Following the common settings for this database, we chose one half of the images for training and the remaining samples were used for testing.
We performed ten experiments by randomly selecting the samples for training and testing.
Based on these experiments, the mean recognition accuracies of different approaches are reported in Table \ref{tab:EYaleB}.
The results for Locality-constrained Linear Coding (LLC) \cite{LLC} is directly taken from \cite{LCKSVD}, where the accuracy is computed using  70 local bases.

Similar to Jiang et al. \cite{LCKSVD}, the sparsity threshold for  K-SVD, LC-KSVD1, LC-KSVD2 and D-KSVD was set to 30 in our experiments. 
Larger values of this parameter were found to be ineffective as they mainly resulted in slowing the algorithms without improving the recognition accuracy. 
Furthermore, as in \cite{LCKSVD}, we used $\upsilon = 4.0$ for LC-KSVD1 and LC-KSVD2, whereas $\kappa$ was set to  $2.0$ for LC-KSVD2 and D-KSVD in Eq.~(\ref{eq:LCKSVD}). 
Keeping these parameter values fixed, we optimized for the number of dictionary atoms for each algorithm.
This resulted in selecting 600 atoms for LC-KSVD2, D-KSVD and K-SVD, whereas 500 atoms consistently resulted in the best performance of LC-KSVD1.
This value is set to 570 in \cite{LCKSVD} for all of the four methods.
In all techniques that learn dictionaries, we used the complete training data in the learning process.
Therefore, all training samples were used as dictionary atoms for SRC.
Following \cite{SRC}, we set the residual error tolerance to 0.05 for SRC.
Smaller values of this parameter also resulted in very similar accuracies.
For FDDL, we followed \cite{FDDL} for the optimized parameter settings. 
These settings are the same as those reported for AR database in the original work.
We refer the reader to the original work for the list of the parameters and their exact values.
The results reported in the table are obtained by the Global Classifier (GC) of FDDL, which showed better performance than the Local Classifier (LC).
For the parameter settings of DL-COPAR we followed the original work \cite{DLCOPAR}. 
We fixed 15 atoms for each class and a set of 5 atoms was chosen to learn commonalities of the classes.
The reported results are achieved by LC, that performed better than  GC  in our experiments. 

\begin{table}[t]
  \caption{Recognition accuracy with Random-Face features on the Extended YaleB database \cite{EYaleB}. The computed average time is for classification of a single instance.}
  \centering
\begin{tabular}{| l | c | c |}
  \hline                       
  Method & Accuracy $\%$ & Average Time (ms)  \\
  \hline \hline
  LLC \cite{LLC}  & $90.7$ & - \\
     K-SVD \cite{KSVD} & $93.13\pm0.43$ & $0.37$  \\
  LC-KSVD1\cite{LCKSVD} & $93.59\pm0.54$ & $0.36$ \\
      D-KSVD \cite{DKSVD} & $94.79 \pm0.49$ &$0.38$ \\
             DL-COPAR \cite{DLCOPAR} & $94.83 \pm 0.52$ & $ 32.55$\\
  LC-KSVD2\cite{LCKSVD} & $95.22\pm0.61$ &$0.39$\\
    FDDL \cite{FDDL} & $96.07\pm 0.64$ & $49.59$\\
  SRC \cite{SRC} & $96.32\pm0.85$ &$53.12$\\
  Proposed & ${\bf 97.19 \pm 0.71}$ & $1.23$ \\
  \hline  
\end{tabular}
\label{tab:EYaleB}
\end{table}

It is clear from Table~\ref{tab:EYaleB} that our approach  outperforms the above mentioned approaches in terms of recognition accuracy, with nearly $23\%$ improvement over the error rate of the second best approach. 
Furthermore, the time required by the proposed approach for classifying a single test instance is also very low as compared to  SRC, FDDL and DL-COPAR. 
For the proposed approach, this time is comparable to D-KSVD and LC-KSVD. 
Like these algorithms, the computational efficiency in the classification stage of our approach comes from using the learned multi-class linear classifier to classify the sparse codes of a test instance. 


\subsection{AR Database}
\label{sec:ARData}
This database contains more than 4,000 color face images of 126 people. 
There are 26 images per person taken during two different sessions. 
In comparison to Extended YaleB, the images in AR database have larger variations in terms of facial expressions, disguise and illumination conditions.
Samples from AR database are shown in Fig.~\ref{fig:ARface} for illustration.
We followed a common evaluation protocol in our experiments for this database, in which we used a subset of 2600 images pertaining to 50 males and 50 female subjects.
For each subject, we randomly chose 20 samples for training and the rest for testing. 
The $165\times120$ pixel images were projected onto a 540-dimensional vector with the help of a random projection matrix, as in Section \ref{sec:EYB}.
We report the average recognition accuracy of our experiments in Table~\ref{tab:AR}, which also includes the accuracy of LLC \cite{LLC} reported in~\cite{LCKSVD}. 
The mean values reported in the table are based on ten experiments. 

In our experiments, we set the sparsity threshold for K-SVD, LC-KSVD1, LC-KSVD2 and D-KSVD to 50 as compared to 10 and 30 which was used in \cite{DKSVD} and \cite{LCKSVD}, respectively.
Furthermore, the dictionary size for K-SVD, LC-KSVD2 and D-KSVD was set to 1500 atoms, whereas the dictionary size for LC-KSVD1 was set to 750.
These large values (compared to 500 used in \cite{DKSVD}, \cite{LCKSVD}) resulted in better accuracies at the expense of more computation.
However, the classification time per test instance remained reasonably small.
In Table~\ref{tab:AR}, we also include the results of  LC-KSVD1, LC-KSVD2 and D-KSVD using the parameter values proposed in the original works.
These results are distinguished with the $\ddagger$ sign. 
For FDDL and DL-COPAR we used the same parameter settings  as in Section \ref{sec:EYB}. 
The reported results are for GC and LC for FDDL and DL-COPAR, respectively. 
For SRC we set the residual error tolerance to $10^{-6}$.
This small value gave the best results.
  
\begin{table}[t]
  \caption{Recognition accuracy with Random-Face features on the AR database \cite{AR}. The computed time is for classifying a single instance. The $\ddagger$ sign denotes the results using the parameter settings reported in the original works.}
  \centering
\begin{tabular}{| l | c | c |}
  \hline                       
  Method & Accuracy $\%$ & Average Time (ms)  \\
  \hline \hline
  LLC \cite{LLC} & 88.7 & - \\
    DL-COPAR \cite{DLCOPAR} & $93.23 \pm 1.71$ &39.80 \\
  LC-KSVD1\cite{LCKSVD} & $93.48\pm 1.13$ & 0.98 \\
     LC-KSVD1$\ddagger$ & $87.48\pm 1.19$ & 0.37 \\
       K-SVD \cite{KSVD} & $94.13\pm1.20$ &  0.99 \\
      LC-KSVD2 \cite{LCKSVD} & $95.33\pm1.24$ & 1.01 \\
  LC-KSVD2$\ddagger$ & $88.35\pm1.33$ & 0.41 \\
       D-KSVD \cite{DKSVD} & $95.47\pm1.50$ & 1.01  \\
  D-KSVD$\ddagger$  & $88.29\pm1.38$ & 0.38  \\
    FDDL \cite{FDDL} & $96.22 \pm 1.03$ & 50.03 \\
  SRC \cite{SRC} & $96.65 \pm 1.37$ & 62.86  \\
  Proposed & ${\bf 97.41\pm1.04}$ & 1.27\\
  \hline  
\end{tabular}
\label{tab:AR}
\end{table}

From Table~\ref{tab:AR}, we can see that the proposed approach performs better than the existing approaches in terms of accuracy. 
The recognition accuracies of SRC and FDDL are fairly close to our approach however, these algorithms require large amount of time for  classification.
This fact compromises their practicality. 
In contrast, the proposed approach shows high recognition accuracy (i.e. $22\%$ reduction in the error rate as compared to SRC) with less than 1.5~ms required for classifying a test instance.
The relative difference between the classification time of the proposed approach and the existing approaches remains similar in the experiments below. 
Therefore, we do not explicitly note these timings for all of the approaches in these experiments.

\begin{figure}[t]
        \centering
        \begin{subfigure}[b]{0.4\textwidth}
                \includegraphics[width=\textwidth]{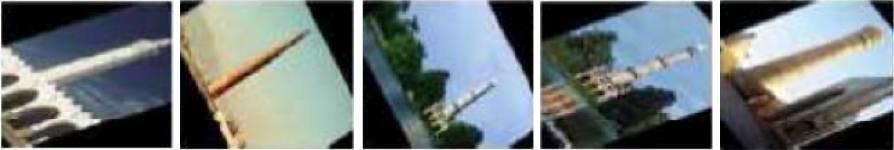}
                \caption{Minaret}
                \label{fig:Minar}
        \end{subfigure}\\%
        \begin{subfigure}[b]{0.4\textwidth}
                \includegraphics[width=\textwidth]{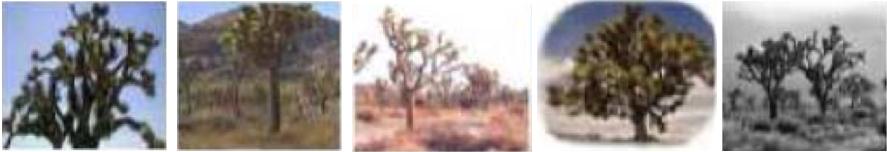}
                \caption{Tree }
                \label{fig:Tree}
        \end{subfigure}\\
        \begin{subfigure}[b]{0.4\textwidth}
                \includegraphics[width=\textwidth]{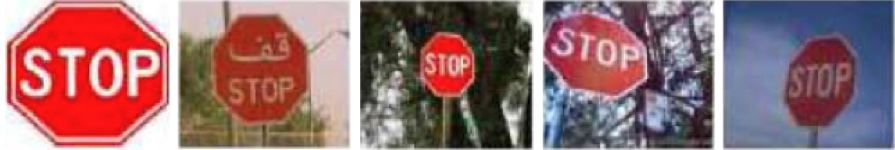}
                \caption{Stop sign }
                \label{fig:Stop}
        \end{subfigure}        
        \caption{Examples from Caltech-101 database \cite{C101}. The proposed approach achieves 100\% accuracy on these classes.}
        \label{fig:C101}
\end{figure}
\begin{table}[t]
  \caption{Classification results using Spatial Pyramid Features on the Caltech-101 dataset \cite{C101}. }
  \centering
\begin{tabular}{| l | c | c | c | c | c | c |}
  \hline                       
  Total training samples & 5  & 10 & 15 & 20 & 25 & 30 \\
  \hline \hline
Zhang et al. \cite{Malik} & 46.6 & 55.8 & 59.1 & 62.0 & - & 66.20\\
Lazebnik et al. \cite{Lazebnik} & - & - & 56.4 & - & - & 64.6\\
Griffin et al. \cite{Griffin} & 44.2 &54.5 &59.0 & 63.3 & 65.8 & 67.6\\
Wang et al. \cite{LLC} & 51.1 & 59.8 & 65.4 & 67.7 & 70.2 & 73.4 \\
SRC \cite{SRC} & 49.9 & 60.1 & 65.0 & 67.5 & 69.3 & 70.9 \\
DL-COPAR \cite{DLCOPAR}& 49.7& 58.9 & 65.2 & 69.1 & 71.0 & 72.9\\
K-SVD \cite{KSVD} 		&51.2 &  59.1 & 64.9 & 68.7 & 71.0 & 72.3 \\
FDDL \cite{FDDL} & 52.1 & 59.8 & 66.2 & 68.9 & 71.3 & 73.1\\
D-KSVD \cite{DKSVD}& 52.1 & 60.8 & 66.1 & 69.6 & 70.8 & 73.1 \\
LC-KSVD1 \cite{LCKSVD} & 53.1 & 61.2 & 66.3 & 69.8 & 71.9 & 73.5 \\
LC-KSVD2 \cite{LCKSVD} & 53.8 & 62.8 & 67.3 & 70.4 & 72.6 & 73.9 \\
Proposed & {\bf 53.9} & {\bf 63.1} & {\bf 67.7} & {\bf 70.9} & {\bf 73.2} & {\bf 74.6} \\
  \hline  
\end{tabular}
\label{tab:C101}
\end{table}

\subsection{Caltech-101}
The Caltech-101 database \cite{C101} comprises $9,144$ samples from 102 classes. 
Among these, there are 101 object classes (e.g. minarets, trees, signs) and one  ``background'' class. 
The number of samples per class varies from 31 to 800, and the images within a given class have significant shape variations, as can be seen in Fig.~\ref{fig:C101}.
To use the database, first the SIFT descriptors \cite{SIFT} were extracted from $16\times16$ image patches, which were densely sampled with a 6-pixels step size for the grid. 
Then, based on the extracted features, spatial pyramid features \cite{Lazebnik} were extracted with $2^l \times 2^l$ grids, where $l = 0,1,2$. 
The codebook for the spatial pyramid was trained using $k$-means with $k = 1024$. 
Then, the dimension of a spatial pyramid feature was reduced to $3000$ using PCA.
Following the common experimental protocol, we selected 5, 10, 15, 20, 25 and 30 instances for training the dictionary and the remaining instances were used in testing, in our six different experiments.
Each experiment was repeated ten times with random selection of train and test data. 
The mean accuracies of these experiments are reported in Table~\ref{tab:C101}.

For this dataset, we set the number of dictionary atoms used by K-SVD, LC-KSVD1, LC-KSVD2 and D-KSVD to the number of training examples available.
This resulted in the best performance of these algorithms. 
The sparsity level was also set to 50 and $\upsilon$ and $\kappa$ were set to $0.001$.
Jiang et al.~\cite{LCKSVD} also suggested the same parameter settings.
For SRC, the error tolerance of $10^{-6}$ gave the best results in our experiments. 
We used the parameter settings for object categorization given in \cite{FDDL} for FDDL.
For DL-COPAR, the selected number of class-specific atoms were kept the same as the number of training instances per class, whereas the number of shared atoms were fixed to 314, as in the original work \cite{DLCOPAR}.
For this database GC performed better than LC for DL-COPAR in our experiments.

From Table~\ref{tab:C101}, it is clear that the proposed approach consistently outperforms the competing approaches. 
For some cases the accuracy of LC-KSVD2 is very close to the proposed approach, however with the increasing number of training instances the difference between the results increases in favor of the proposed approach.
This is an expected phenomenon since more training samples result in more precise posterior distributions in Bayesian settings.
Here, it is also worth mentioning that being Bayesian, the proposed approach is inherently an online technique.
This means, in our approach, the computed posterior distributions can be used as prior distributions for further inference if more training data is available.
Moreover, our approach is able to handle a batch of large training data more efficiently than LC-KSVD \cite{LCKSVD} and D-KSVD~\cite{DKSVD}. 
This can be verified by comparing the training time of the approaches in Table~\ref{tab:C101Time}.
The timings are given for complete training and testing durations for Caltech-101 database, where we used a batch of 30 images per class for training and the remaining images were used for testing.
We note that, like all the other approaches, good initialization (using the procedure presented in Section~\ref{sec:Init}) also contributes towards the computational efficiency of our approach.
The training time in the table also includes the initialization time for all the approaches.
Note that the testing time of the proposed approach is very similar  to those of the other approaches in Table~\ref{tab:C101Time}.

 \begin{table}[t]
  \caption{Computation time for training and testing on Caltech-101 database}
  \centering
\begin{tabular}{| l | c | c |}
  \hline
  Method & Training (sec) & Testing (sec)\\
  \hline
  \hline
  Proposed & 1474 & 19.96\\
  D-KSVD\cite{DKSVD} & 3196  & 19.90 \\
  LC-KSVD1\cite{LCKSVD} & 5434  & 19.65 \\
  LC-KSVD2\cite{LCKSVD} & 5434  & 19.92 \\
 \hline  
\end{tabular}
\label{tab:C101Time}
\end{table}

\subsection{Fifteen Scene Category}
\label{sec:FSC}

The Fifteen Scene Category dataset \cite{Scene15} has 200 to 400 images per category for fifteen different kinds of scenes.
The scenes include images from kitchens, living rooms and country sides etc.
In our experiments, we used the Spatial Pyramid Features of the images, which have been made public by Jiang et al.~\cite{LCKSVD}.
In this data, each feature descriptor is a 3000-dimensional vector.
Using these features, we performed experiments by randomly selecting 100 training instances per class and considering the remaining as the test instances.

Classification accuracy of the proposed approach is compared with the existing approaches in Table~\ref{tab:Scene15}.
The reported mean values are computed over ten experiments. 
We set the error tolerance for SRC to $10^{-6}$ and used the parameter settings suggested by Jiang et al.~\cite{LCKSVD} for LC-KSVD1, LC-KSVD2 and D-KSVD.
Parameters of DL-COPAR were set as suggested in the original work~\cite{DLCOPAR} for the same database. 
The reported results are obtained by LC for DL-COPAR.
Again, the proposed approach shows more accurate results than the existing approaches.
The accuracy of the proposed approach is $1.66\%$ more than LC-KSVD2 on the used dataset.

\begin{figure}[t] 
   \centering
   \includegraphics[width=3.5in]{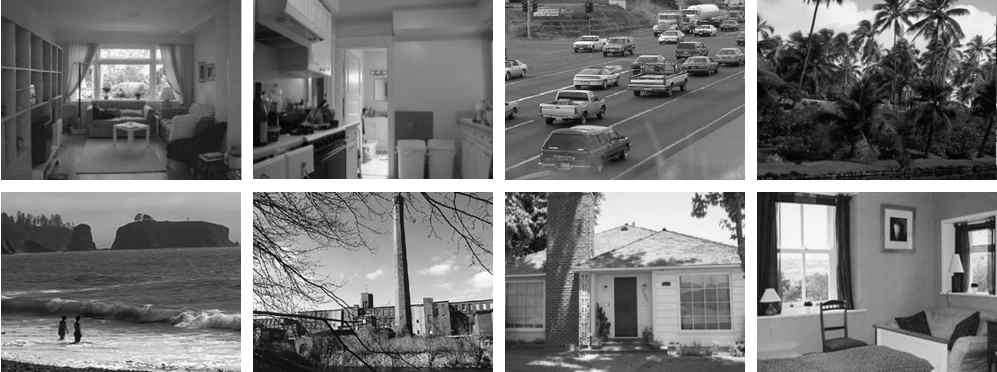} 
   \caption{Examples images from eight different categories in Fifteen Scene Categories dataset \cite{Scene15}.}
   \label{fig:Scene15}
\end{figure}
\begin{table}[t]
  \caption{Classification accuracy on Fifteen Scene Category dataset \cite{Scene15} using Spatial Pyramid Features.}
  \centering
\begin{tabular}{| l | c |}
  \hline                       
  Method & Accuracy $\%$   \\
  \hline \hline
     K-SVD \cite{KSVD} & $93.60\pm0.14$   \\
  LC-KSVD1\cite{LCKSVD} & $94.05\pm0.17$  \\
      D-KSVD \cite{DKSVD} & $96.11 \pm0.12$  \\
        SRC \cite{SRC} & $96.21\pm0.09$ \\
     DL-COPAR \cite{DLCOPAR} & $96.91 \pm 0.22$\\
  LC-KSVD2\cite{LCKSVD} & $97.01\pm0.23$ \\
  Proposed & ${\bf 98.67 \pm 0.19}$  \\
  \hline  
\end{tabular}
\label{tab:Scene15}
\end{table}

\subsection{UCF Sports Action}
\begin{figure}[t] 
   \centering
   \includegraphics[width=3.5in, height = 1.3in]{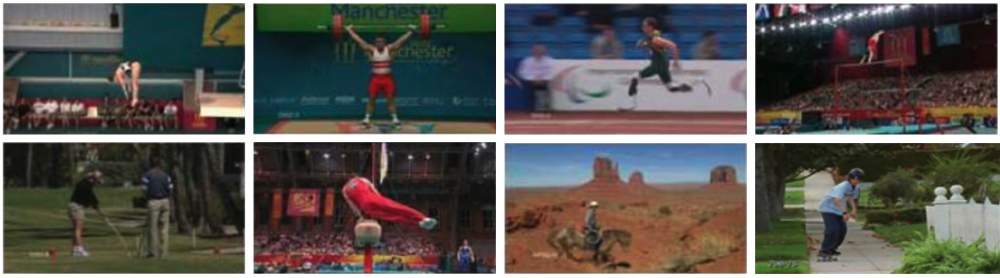} 
   \caption{Examples from UCF Sports action dataset \cite{UCF}. }
   \label{fig:UCF}
\end{figure}
\begin{table}[t]
  \caption{Classification rates on UCF Sports Action dataset~\cite{UCF}}
  \centering
\begin{tabular}{| l | c || l | c |}
\hline
Method & Accuracy $\%$ & Method & Accuracy $\%$\\
  \hline\hline
 Qiu et al. \cite{A1} 				& 83.6 		& LC-KSVD2 \cite{LCKSVD} & 91.5  \\
 D-KSVD \cite{DKSVD} 			& 89.1			& DLSI \cite{DLSI} & 92.1  \\
 LC-KSVD1 \cite{LCKSVD} 		& 89.6  		& SRC \cite{SRC} & 92.7\\
 DL-COPAR \cite{DLCOPAR} 	& 90.7			& FDDL \cite{FDDL} & 93.6\\
Sadanand \cite{Sadanand} 		& 90.7  		& LDL \cite{LDL} & 95.0\\
Proposed & {\bf 95.1}  & &\\
 \hline
  \end{tabular}
  \label{tab:UCF}
  \end{table} 
This database comprises video sequences that are collected from different broadcast sports channels (e.g. ESPN and BBC) \cite{UCF}. 
The videos contain 10 categories of sports actions that include: kicking, golfing, diving, horse riding, skateboarding, running, swinging, swinging highbar, lifting and walking.
Examples from this dataset are shown in Fig.~\ref{fig:UCF}. 
Under the common evaluation protocol we performed fivefold cross validation over the dataset,  where four folds are used in training and the remaining one is used for testing.
Results, computed as the average of the five experiments, are summarized in Table \ref{tab:UCF}.
For D-KSVD, LC-KSVD1 and LC-KSVD2 we followed \cite{LCKSVD} for the parameter settings. 
Again, the value of $10^{-6}$ (along with similar small values) resulted in the best accuracies for SRC.

In the Table, the results for some specific action recognition methods are also included, for instance, Qui et al.~\cite{A1} and action back feature with SVM~\cite{Sadanand}.
These results are taken directly from \cite{LDL} along the results of DLSI~\cite{DLSI}, DL-COPAR \cite{DLCOPAR} and FDDL~\cite{FDDL}\footnote{The results of DL-COPAR \cite{DLCOPAR} and FDDL~\cite{FDDL} are taken directly from the literature because the optimized parameter values for these algorithms are not previously reported for this dataset. Our parameter optimization did not outperform the reported accuracies.}.
Following \cite{Sadanand}, we also performed leave-one-out cross validation on this database for the proposed approach.
Our approach achieves $95.7\%$ accuracy under this protocol, which is $0.7\%$ better than the state-of-the-art results claimed in \cite{Sadanand}. 

\section{Discussion}
\label{sec:D}
\begin{figure*}
\includegraphics[width=\textwidth]{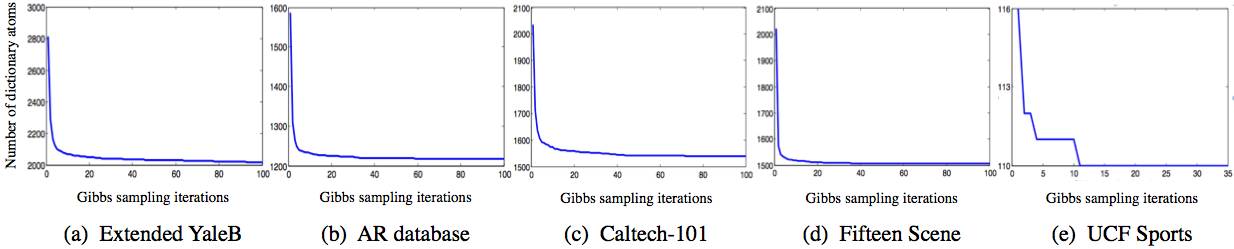}
        \caption{Size of the inferred dictionary, i.e. $|\mathcal K|$, as a function of the Gibbs sampling iterations. Each plot represents a complete training process for a given dataset.} 
        \label{fig:atoms}
\end{figure*}

In our experiments, we chose the values of $K, a_o$ and $b_o$ in light of the theoretical results presented in Section~\ref{sec:I}.
By setting $K > N$ we make sure that $K$ is very large.
The results mainly remain insensitive to other similar large values of this parameter.
The chosen values of $a_o$ and $b_o$ ensure that $0 < a_o, b_o < |\mathcal I_c |$. 
We used large values for $\lambda_{\epsilon_o}$ in our experiments as this parameter represents the precision of the white noise distribution in the samples.
The datasets used in our experiments are mainly clean in terms of white noise.
Therefore, we achieved the best performance with $\lambda_{\epsilon_o} \geq 10^6$. 
In the case of noisy data, this parameter value can be adjusted accordingly.
For UCF sports action dataset $\lambda_{\epsilon_o} = 10^9$ gave the best results because less number of training samples were available per class.
It should be noted that the value of $\lambda_{\epsilon}$  increases as a result of Bayesian inference with the availability of  more clean training samples.
Therefore, we adjusted the precision parameter of the prior distribution to a larger value for UCF dataset.  
Among the other parameters, $c_o$ to $f_o$ were fixed to $10^{-6}$.
Similar small non-negative values can also be used without affecting the results.
This fact can be easily verified by noticing the large values of the other variables involved in equations (\ref{eq:12}) and (\ref{eq:13}), where these parameters are used.
With the above mentioned parameter settings and the initialization procedure presented in Section~\ref{sec:Init}, the Gibbs sampling process converges quickly to the desired distributions and the correct number of dictionary atoms, i.e. $|\mathcal K|$.
In Fig.~\ref{fig:atoms}, we plot the value of $|\mathcal K|$ as a function of Gibbs sampling iterations during dictionary training.
Each plot represent a complete training process for one dataset. 
It can be easily seen that the first 10 iterations of the Gibbs sampling process were enough to infer the correct size of the dictionary.
However, It should be mentioned that this fast convergence also owes to the initialization process adopted in this work.
In  our experiments, while  sparse coding a test instance over the learned dictionary, we consistently used the sparsity threshold of $50$ for all the datasets except for the UCF \cite{UCF}, for which this parameter was set to $40$ because of the smaller dictionary resulting from less training samples. 
In all the experiments, these values were also kept the same for K-SVD, LC-KSVD1, LC-KSVD2 and D-KSVD for fair comparisons.

\section{Conclusion}
\label{sec:Conc}
We proposed a non-parametric Bayesian approach for learning discriminative dictionaries for sparse representation of data.
The proposed approach employs a Beta process to infer a discriminative dictionary and sets of Bernoulli distributions associating the dictionary atoms to the class labels of the training data. 
The said association is adaptively built during Bayesian inference and it signifies the selection probabilities of  dictionary atoms in the expansion of class-specific data.
The inference process also results in computing the correct size of the dictionary.
For learning the discriminative dictionary, we presented a  hierarchical Bayesian model and the corresponding inference equations for Gibbs sampling.
The proposed model is also exploited in learning a linear classifier that finally classifies the sparse codes of a test instance that are learned using the inferred discriminative dictionary. 
The proposed approach is evaluated for classification using five different databases of human face, human action, scene category and object images. 
Comparisons with state-of-the-art discriminative sparse representation approaches show that the proposed Bayesian approach consistently outperforms these approaches and has computational efficiency close to the most efficient approach.
 
Whereas its effectiveness in terms of accuracy and computation is experimentally proven in this work, there are also other key advantages that make our Bayesian approach to discriminative sparse representation much more appealing than the existing optimization based approaches.
Firstly, the Bayesian framework allows us to learn an ensemble of discriminative dictionaries in the form of probability distributions instead of the point estimates that are learned by the optimization based approaches. 
Secondly, it provides a principled approach to estimate the required dictionary size and we can associate the dictionary atoms and the class labels in a physically meaningful manner.
Thirdly, the Bayesian framework makes our approach inherently an online technique.
Furthermore, the Bayesian framework also provides an opportunity of using domain/class-specific prior knowledge in our approach in a principled manner.
This can prove beneficial in many applications.
For instance, while classifying the spectral signatures of minerals on pixel and sub-pixel level in remote-sensing hyperspectral images, the relative smoothness of spectral signatures~\cite{myTGRS} can be incorporated in the inferred discriminative bases.
For this purpose, Gaussian Processes~\cite{GP} can be used as a base measure for the Beta Process.
Adapting the proposed approach for remote-sensing hyperspectral image classification is also our future research direction.


%




\ifCLASSOPTIONcompsoc
  \section*{Acknowledgments}
\else
  \section*{Acknowledgment}
\fi

This research was supported by ARC Grant DP110102399.

\ifCLASSOPTIONcaptionsoff
  \newpage
\fi



%

%

\begin{IEEEbiography}[{\includegraphics[width=1in,height=1.25in,clip,keepaspectratio]{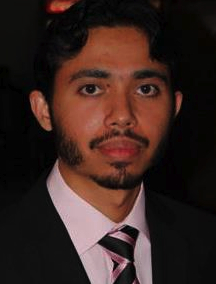}}]{Naveed Akhtar}
received BE degree with distinction in Avionics from the College of Aeronautical Engineering, National University of Sciences and Technology (NUST), Pakistan, in 2007 and M.Sc. degree with distinction in Autonomous Systems from Hochschule Bonn-Rhein-Sieg (HBRS), Sankt Augustin, Germany, in 2012.
He is currently working toward the Ph.D. degree at The University of Western Australia (UWA) under the supervision of Dr. Faisal Shafait and Prof. Ajmal Mian.
He was a recipient of competitive scholarship for higher studies by Higher Education Commission, Pakistan in 2009 and currently he is a recipient of SIRF scholarship at UWA.
He has served as a Research Assistant at Research Institute for Microwaves and Millimeter-waves Studies, NUST, Pakistan, from 2007 to 2009 and as a Research Associate at the Department of Computer Science at HBRS, Germany in 2012. 
His current research is focused on sparse representation based image analysis.  
\end{IEEEbiography}

\begin{IEEEbiography}[{\includegraphics[width=1in,height=1.25in,clip,keepaspectratio]{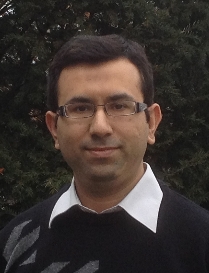}}]{Faisal Shafait}
is working as an Assistant Professor in the Computer Science and Software Engineering Department at The University of Western Australia. Formerly, he was a Senior Researcher at the German Research Center for Artificial Intelligence (DFKI), Germany and a visiting researcher at Google, California. He received his Ph.D. in computer engineering with the highest distinction from Kaiserslautern University of Technology, Germany in 2008. His research interests include machine learning and pattern recognition with a special emphasis on applications in document image analysis. He has co-authored over 80 publications in international peer-reviewed conferences and journals in this area. He is an Editorial Board member of the International Journal on Document Analysis and Recognition (IJDAR), and a Program Committee member of leading document analysis conferences including ICDAR, DAS, and ICFHR. He is also serving on the Leadership Board of IAPR?s Technical Committee on Computational Forensics (TC-6).
\end{IEEEbiography}
\begin{IEEEbiography}[{\includegraphics[width=1in,height=1.25in,clip,keepaspectratio]{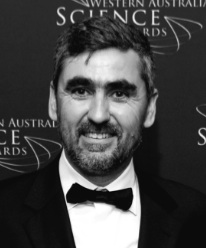}}]{Ajmal Mian}
received the BE degree in Avionics from the College of Aeronautical Engineering, Nadirshaw Edulji Dinshaw (NED) University, Pakistan, in 1993, the MS degree in Information Security from the National University of Sciences and Technology, Pakistan, in 2003, and the PhD degree in computer science with distinction from The University of Western Australia in 2006. He received the Australasian Distinguished Doctoral Dissertation Award from Computing Research and Education Association of Australia (CORE) in 2007. He received two prestigious fellowships, the Australian Postdoctoral Fellowship in 2008 and the Australian Research Fellowship in 2011. He was named the West Australian Early Career Scientist of the Year 2012. He has secured four national competitive research grants and is currently a Research Professor in the School of Computer Science and Software Engineering, The University of Western Australia. His research interests include computer vision, pattern recognition, machine learning, multimodal biometrics, and hyperspectral image analysis.
\end{IEEEbiography}



\end{document}